\documentclass{IEEEtran}
\usepackage{cite}
\usepackage{amsmath,amssymb,amsfonts}
\usepackage{algorithmic}
\usepackage{graphicx}
\usepackage{textcomp}
\usepackage{comment}
\usepackage{subfigure}
\usepackage{multirow}
\usepackage{caption}
\usepackage{color, soul}
\usepackage{booktabs}

\long\def\/*#1*/{}

\def\BibTeX{{\rm B\kern-.05em{\sc i\kern-.025em b}\kern-.08em
    T\kern-.1667em\lower.7ex\hbox{E}\kern-.125emX}}
    
\begin{document}



\title{Enhance Gender and Identity Preservation in Face Aging Simulation for Infants and Toddlers}

\author{\IEEEauthorblockN{Yao Xiao\IEEEauthorrefmark{1},
Yijun Zhao\IEEEauthorrefmark{1}
}
\thanks{\copyright 2020 This work has been submitted to the IEEE for possible publication. Copyright may be transferred without notice, after which this version may no longer be accessible.}

\vspace{4mm}

\IEEEauthorblockA{\IEEEauthorrefmark{1} Computer and Information Science Department, Fordham University, New York, NY, USA
}

}


\maketitle



\begin{abstract}
Realistic age-progressed photos provide invaluable biometric information in a wide range of applications. In recent years, deep learning-based approaches have made remarkable progress in modeling the aging process of the human face. Nevertheless, it remains a challenging task to generate accurate age-progressed faces from infant or toddler photos. In particular, the lack of visually detectable gender characteristics and the drastic appearance changes in early life contribute to the difficulty of the task. We propose a new deep learning method inspired by the successful Conditional Adversarial Autoencoder (CAAE, 2017) model. In our approach, we extend the CAAE architecture to 1) incorporate gender information, and 2) augment the model's overall architecture with an identity-preserving component based on facial features. We trained our model using the publicly available UTKFace dataset and evaluated our model by simulating up to 100 years of aging on 1,156 male and 1,207 female infant and toddler face photos. Compared to the CAAE approach, our new model demonstrates noticeable visual improvements. Quantitatively, our model exhibits an overall gain of 77.0\% (male) and 13.8\% (female) in gender fidelity measured by a gender classifier for the simulated photos across the age spectrum. Our model also demonstrates a 22.4\% gain in identity preservation measured by a facial recognition neural network.
\end{abstract}



\section{Introduction}

Face progression, also known as face aging, is the process of rendering face images to different age groups to simulate an identity-preserving aging effect. Generating accurate prediction images from a reference photo facilitates many practical applications to help create a safe and secure society. For example, the quality of age-progressed photos plays a critical role in finding long lost children, identifying fugitives, as well as developing age-invariant security systems. Furthermore, being described as "half art and half science", face progression also benefits the entertainment, cosmetology, and biometrics industries. 

In recent years, deep learning-based approaches have made remarkable progress in modeling the aging process of the human face~\cite{liu2017face, radford2015unsupervised, zhu2017unpaired, antipov2017face, zhang2017age, song2018binary}.  Nevertheless, generating accurate age-progressed faces is still a challenging task. This is particularly true when the original face photos are of young children. For example, infants and toddlers often lack salient gender characteristics. Thus, a machine learning model may guess a child's sex incorrectly and produce an inaccurate simulation. Furthermore, even with a correct guess, such models can fail to capture the drastic physical changes which occur during early childhood and adolescence. As a result, predicting future appearance based on a two-year-old face photo is significantly harder than extrapolating from a 20-year-old one.

In this paper, we introduce a new deep learning model to address the two challenges mentioned above. Our approach is
inspired by the highly successful Conditional Adversarial Autoencoder (CAAE)~\cite{zhang2017age} model. First, we incorporate additional gender information into the latent vector of the CAAE architecture. We evaluate the gender consistency of the generated face images using a gender classifier based on a deep neural network model \cite{levi2015age}.
Our study suggests that the CAAE model is prone to generate female faces when sex characteristics are not evident on the original photo. Providing additional gender information is essential in helping the model make proper adjustments.





The second contribution of our study is to enhance identity preservation during the aging simulation for young faces. Before entering early adulthood, the human body undergoes rapid physical growth and puberty development. These physiological changes (e.g., height, weight, and voice changes) are accompanied by substantial changes in facial features. For example, compared to the archetypal adult face, an infant's face tends to have bigger eyes, shorter and flatter eyebrows, and a smaller and turned-up nose. As a result, it is challenging to preserve identity from baby photos in simulating the aging process. To address this issue, we leverage the popular VGG\cite{simonyan2014very} architecture to maximize the similarity of high-level facial features between the original and simulated faces. 
Compared to the baseline CAAE model, our approach demonstrates significant improvement in identity preservation as measured by a face recognition deep neural network developed by Schroff \textit{et al}. \cite{schroff2015facenet}.

The rest of the paper is organized as follows. We present a brief literature survey of related work in Section \ref{sec:related}. 
We then illustrate our proposed model in Section \ref{sec:methods}. We present and discuss our experimental methods and results in Section \ref{sec:results}. Lastly, we conclude in Section \ref{sec:conclusion}.

\section{Related Work}
\label{sec:related}
\subsection{Face Age Progression/Regression}

Human face progression/regression has always been an active area of research due to its broad applications across various disciplines \cite{shu2016age, ramanathan2009computational, ramanathan2009age, fu2010age }. In its early stages, face progression techniques simulated biomechanical aspects of human skin (e.g., anisotropy, visco-elasticity, etc.) to artificially produce wrinkles, creases, and folds \cite{boissieux2000simulation}. These simulations accounted for many specific aging factors, including the facial muscles~\cite{suo2012concatenational, berg2003aging}, cranium~\cite{todd1980perception}, facial skin~\cite{bando2002simple, boissieux2000simulation, lee1999cloning, wu1999skin}, depth of wrinkles~\cite{ramanathan2008modeling, suo2009compositional}, facial structure~\cite{ramanathan2006modeling, lanitis2002toward}, etc. While successful, these methods involved parameter tweaking and domain-specific knowledge. As the state-of-the-arts transitioned to image data driven approaches, methods which modeled face aging as a mapping from young to old faces became more common. For example, Park \textit{et al}. \cite{park2010age} used sequential face data to approximate the aging process as a Markov process. Shu \textit{et al}. presented an efficient and effective Kinship-Guided Age Progression (KinGAP) approach \cite{shu2016kinship}, which can automatically generate personalized aging images with guidance of the senior kinship face. One limitation of these methods is the difficulty of collecting a large dataset with chronologically sequential faces for each individual. 


Another data-based approach is the prototype-based method, which first divides all the faces into different age groups. The technique then estimates the average face within a predefined age group denoted as the prototype. Differences between prototypes form the axes, along which faces can smoothly transition across the aging spectrum. Prototype-based methods are simple, straightforward, and fast. However, since the generated faces are averaged, high frequency details such as wrinkles are not accurately captured. To address this issue, researchers developed models \cite{tang2017personalized, yang2016face, wang2016category}, which captured person-specific facial features via sparse representation. 

\subsection{Deep Learning-based Approaches}
In recent years, deep learning techniques have made remarkable progress in the quality of face progression/regression using large datasets of raw face images. Wang \textit{et al}.\cite{wang2016recurrent} 
employed a recurrent neural network \cite{schuster1997bidirectional} model to capture the common transition patterns among all age groups. Leveraging the slowly evolving intermediate images, their model offers more realistic and smoothly progressing faces than other non-temporal models. The limitation of their approach is that it can only perform one directional (i.e., from young to old) simulation.

Another class of popular deep learning-based methods is derived from the novel generative adversarial networks (GANs)\cite{goodfellow2014generative} framework, in which a model trains a generator and a discriminator simultaneously. While the generator strives to generate realistic faces, the discriminator acts as the adversary to distinguish fake (i.e., generated) vs. real faces. At equilibrium, the generator is forced to produce photo-realistic faces to fool the discriminator. Generative approaches facilitate both face progression and rejuvenation on a face photo of any age. 

One variant of the GAN architecture is the conditional generative adversarial network (C-GAN) \cite{mirza2014conditional}. C-GANs models provide the generator with additional information to control the scope (e.g., a specific age) of the generated images. These models \cite{liu2017face, shu2015personalized, wang2016recurrent} are effective, but they require paired images in training their models, i.e., pictures of the same person across the entire age spectrum, which is difficult to collect.
Recent work by Zhang \textit{et al}. \cite{zhang2017age} overcomes this difficulty by combining the concepts of C-GAN and autoencoder\cite{hinton2006reducing}. Their conditional adversarial autoencoder (CAAE) model embeds age-specific information in the latent vector of the autoencoder. Additionally, the authors employed an extra discriminator to impose a uniform prior on the latent distribution of the input data. We give a more detailed introduction to the CAAE model in Section \ref{sec:caae} because our model is an extension to its architecture. 

\subsection{Our Contribution}
Most of the existing studies focus on the photo-realistic quality of the generated faces, including richer face texture (e.g., wrinkles), smooth transitions, and lesser ghosting effects. Evaluations are typically conducted using visual examinations centered on adult and senior faces. 
In our study, we focus on face aging simulation on baby photos in which the gender information is not detectable and the facial features undergo drastic changes over the years. To the best of our knowledge, there is no prior work addressing this task that is both challenging and of significant practical value. We also establish rigorous quantitative metrics to evaluate the efficacy of our approach.

\begin{figure*}[!t]
\centering
\includegraphics[trim = 0 50 0 0, clip,scale = 0.6]{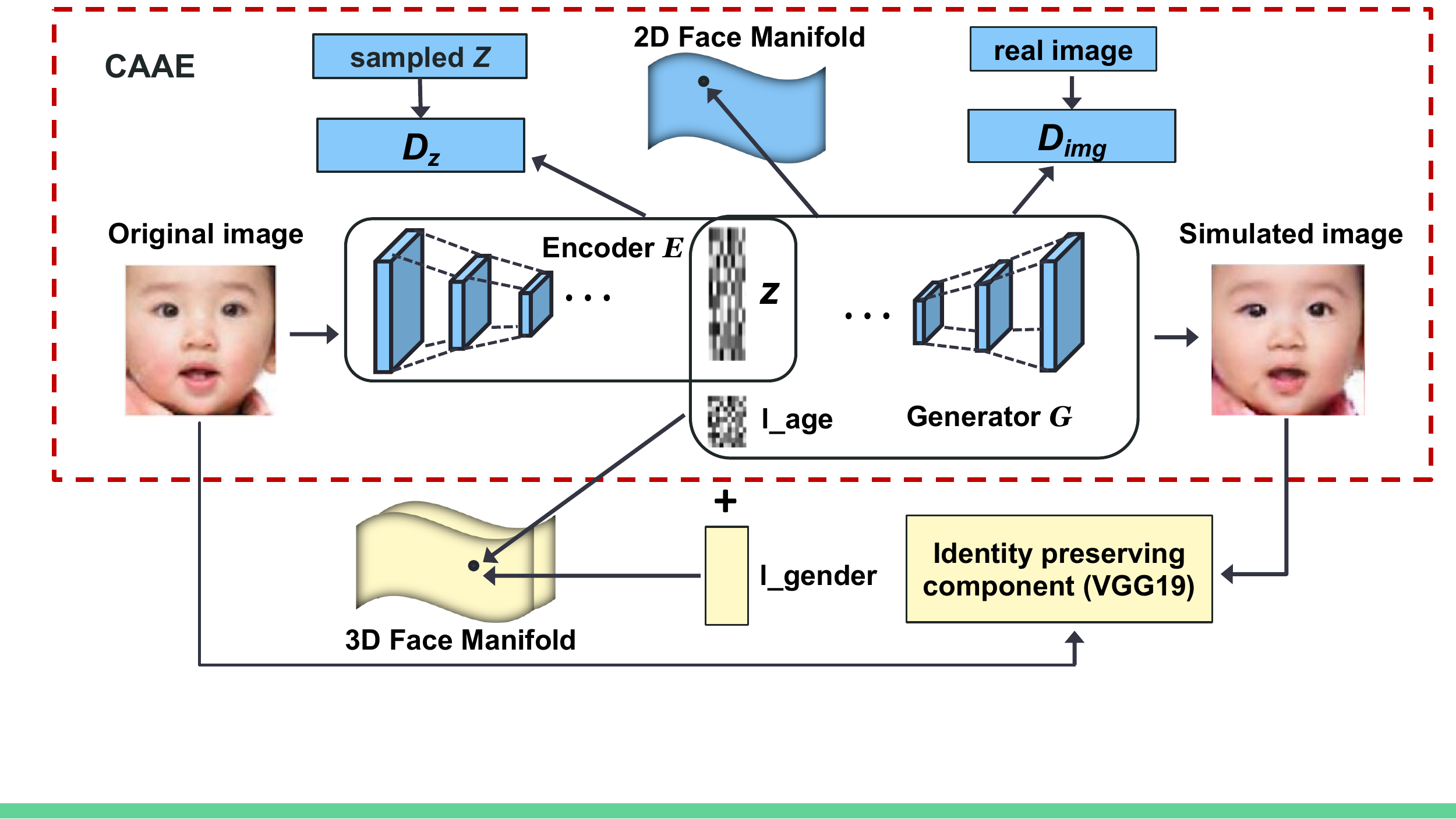}
\caption{Architecture of our Proposed Model}
\label{fig:model}
\end{figure*}

\section{Methods}
\label{sec:methods}
In this section, we first give a brief introduction to the conditional adversarial autoencoder (CAAE) model \cite{zhang2017age}.
We then present our proposed approach which extends the CAAE architecture with gender and identity preserving components.

\subsection{Conditional Adversarial Autoencoder} \label{sec:caae}
As illustrated within the dashed bounding box in Fig. \ref{fig:model}, a CAAE \cite{zhang2017age} model consists of three separate neural networks: a convolutional autoencoder (AE, i.e., Encoder + Generator) and two adversarial discriminators (i.e., $D_{z}$ and $D_{img}$). For the AE component, the encoder ($E$) maps the input face to a vector $z$ which can be interpreted as the personality of a particular face input $x$. The generator ($G$) regenerates the input photo based on the latent vector $z$ and an additional age label ($l$). Thus, a concatenated latent vector $[z, l]$ serves as the input to the generator. Incorporating specific age information into the latent space facilitates clustering face images by age groups. Consequently, the model produces smooth age transitions and forces the output face to be photo-realistic and plausible for a given age.

The second component is the discriminator $D_{z}$ which regularizes the latent space $z$ using a uniform distribution. The uniform prior forces $z$ to evenly populate the latent space without large gaps. Aassuming that images of different personalities and ages form a 2D manifold ($M$) along the two corresponding directions, generating age-progressed photos can be viewed as traversing the $M$ along the age direction. Thus, continuity in the latent space is desirable because it ensures sufficient intermediate data points to guide the model along the traversal. 

The last component $D_{img}$ is an inherited concept from the effective GAN\cite{goodfellow2014generative} methodology. In particular, while the generator strives to produce as realistic as possible photos for a given age, $D_{img}$ acts as an adversary to discriminate between real and generated images. During the training process, $D_{img}$ contributes to the loss function based on how well it detects generated images, while the generator is evaluated based on how well its constructed images can fool the discriminator. Thus, this adversarial process forces the generator to produce high quality images that are indistinguishable from the real ones.    

The overall loss of the CAAE model is the sum of corresponding losses from its three components. Following the same notation as in \cite{zhang2017age}, the objective function can be formulated as follows:
\begin{equation}
\label{eq:caae}
\begin{aligned}
\min_{E,G}&\max_{D_z,D_{img}} \lambda\mathcal{L}(x,G(E(x),l)) + \gamma TV(G(E(x),l))\\
&+\mathbb{E}_{z^*\sim p(z)}[\text{log }D_z(z^*)]\\
&+\mathbb{E}_{x\sim p_{data}(x)}[\text{log }(1 - D_z(E(x)))]\\
&+\mathbb{E}_{x,l\sim p_{data}(x,l)}[\text{log }D_{img}(x,l)]\\
&+\mathbb{E}_{x,l\sim p_{data}(x,l)}[\text{log } (1 - D_{img}(G(E(x),l)))]
\end{aligned}
\end{equation}
where $\mathcal{L}(\cdot, \cdot)$ denotes $L_2$ norm, $p_{data}(x)$ denotes the distribution of the training data, and $z^*\sim p(z)$ denotes the random sampling process from the prior distribution of $z$. The first line of equation (\ref{eq:caae}) is the loss of the CAE component. Specifically, $\mathcal{L}(\cdot, \cdot)$ and $TV(\cdot)$ represent the image reconstruction loss and total variation respectively. The coefficients $\lambda$ and $\gamma$ balance the smoothness and high resolution. The remaining four terms are the standard adversarial loss for discriminators $D_z$ and $D_{img}$ respectively.

\subsection{Our Approach} \label{sec:impr}

Fig. \ref{fig:model} presents the structure of our proposed model. The components outside the dashed box are extensions to the CAAE model. 

\subsubsection{Integrating Gender Information}

One limitation of the CAAE model is that it does not account for gender differences in the input data. While gender can generally be inferred from adult photos, this is not the case with infant or toddler pictures. We demonstrate in our experimental results that the CAAE model tends to age-progress a baby photo toward the female direction. To address this issue, we augment the latent space $z$ with additional gender information. As illustrated in Fig. \ref{fig:model}, the input to our generator $G$ is $[z, l, s]$ where $s \in \{male, female\}$. We can further interpret our model as a 3D extension to the 2D manifold of the CAAE model with gender being the third dimension. 

We evaluate the efficacy of integrating the new gender indicator using simulated aging images constructed from 0-5 years old photos. Besides visual examinations, we quantify the generated images with a gender score measured by a gender classifier. We present our gender classifier and the evaluation results in Section \ref{sec:results}.

\subsubsection{Enhancing Identify Preservation }
\label{sec:identity}

Our second improvement is to leverage a VGG \cite{simonyan2014very} model to preserve each individual's identity in the face aging process. The VGG network was introduced by Simonyan and Zisserman in 2014. 
It employs deep convolutional layers and is one of the most popular architectures for image feature extraction. In our approach, we utilize a pre-trained VGG19 (i.e., 19 convolutional layers) model to minimize the differences in high-level facial features between the input and generated images. Specifically, we apply the VGG19 model to both the input and the generated images and compute the $L_2$ distance between the two feature maps after the last convolutional layer. The difference is added to the total training loss to force the model to maintain the same personalities as the original faces in its aging simulation process.

We evaluate the efficacy of this new identity preservation component using a Face Recognition (FR) score from the FaceNet system\cite{schroff2015facenet} developed by Google in 2015. Specifically, the FaceNet system maps a pair of images to an embedded Euclidean space, and the distance (e.g., $L_2$) in the embedded space measures the similarity of two identities. We present the FR score calculation and the evaluation of the VGG component in Section \ref{sec:results}.

\subsubsection{Objective Function} \label{sec:loss}
We make two modifications to the objective function in Equation (\ref{eq:caae}) to serve the two new components in our model. First, we augment the terms of adversarial loss of $D_{img}$ with additional gender information ($s$). Formally, the last two lines in Equation (\ref{eq:caae}) are adjusted as follows:

\vspace{2mm}
\noindent \hspace{1cm}$\mathbb{E}_{x,l,s\sim p_{data}(x,l,s)}[\text{log }D_{img}(x,l,s)] + \\
\hspace*{1cm} \mathbb{E}_{x,l,s\sim p_{data}(x,l,s)}[\text{log } (1 - D_{img}(G(E(x),l,s)))]\\$

Next we add a new term to Equation (\ref{eq:caae}) representing the feature map differences between the input and simulated images. Formally, the loss can be expressed as:

\vspace{2mm}
\hspace*{1cm} {$\mathcal{L}(FM(x),FM(G(E(x)),l,s))$}
\vspace{2mm}

\noindent where $FM(\cdot)$ denotes the feature map vector of the last convolutional layer after applying VGG19 to an image. $s$ is the gender indicator. All other symbols have been kept the same as in Equation (\ref{eq:caae}).

\section{Experiments}
\label{sec:results}

In this section, we first describe the data we used to conduct our study. We then introduce two quantitative measures to evaluate the quality of gender and identify preservation for the age-progressed photos. Lastly, we present the experimental results of our model in comparison to the baseline CAAE model.

\subsection{Data Acquisition}
We conduct our experiments using the UTKFace dataset\cite{geraldsutkface}, a large-scale face dataset with a long age span (from 0 to 116 years old). The dataset consists of 12,391 male and 11,317 female face images with annotations of age, gender, and ethnicity.  We divided the photos into ten groups (i.e., 0-5, 6-10, 11-15, 16-20, 21-30, 31-40, 41-50, 51-60, 61-70, and >70) 
according to the stages of life in which the human face goes through significant changes. Additionally, the face images are cropped to 228$\times$228 and aligned to make training more tractable.

\subsection{Quantitative Metrics}

Besides visually examining the quality of generated photos, we introduce "gender score" and "FR (Face Recognition) score" to assess the sex and identity consistencies between the original and simulated images.  


\subsubsection{Gender Score} \label{sec:gender_clf}
\label{sec:gs}

 To quantitatively evaluate gender fidelity during the face aging process, we employ a binary gender classifier ($C$) to measure masculinity and femininity in the generated photos. We define the male/female gender score for a given age group as the accuracy when applying $C$ to all of the expected male/female photos in the group. For example, the male gender score for the 16-20-year old group is the accuracy of applying $C$ to those photos in the group that are simulated from male input. Thus, a higher gender score is more desirable because it indicates higher consistency with the expected sex. 
 
 Our gender classifier is based on the architecture of the deep neural network model introduced by Levi and Hassner~\cite{levi2015age} in 2015. We trained our model using a total of 8,659 male and 7,936 female adult instances from the UTKFace dataset with a 70\%, 15\%, 15\% split for model training, validation, and testing respectively. Table \ref{tb:gender_acc} presents the gender scores of each age group in the test data. We observe that our gender classifier has an over 90\% average accuracy for both classes. Furthermore, the model has a more balanced performance in the middle section (21 to 70 years old) of the age spectrum. Since we are interested in the relative performance of various models within a specific class and age group, the imbalanced performance at the two tails is insignificant because the same biases will apply to all models.  
 
  \renewcommand{\arraystretch}{1.4}
\begin{table}[!t] 
\small
\caption{Performance of Gender Classifier $C$}
\label{tb:gender_acc}
\begin{center}
	\begin{tabular}{c|l|l}
	\toprule
	Age Group &  Male & Female \\
	\hline
	$0-5$ & 0.77 (138/179)* & 0.70 (128/183) \\
	$6-10$ & 0.75 (45/60) & 0.96 (75/78) \\
	$11-15$ & 0.72 (29/40) & 0.93 (51/55) \\
	$16-20$ &0.88 (52/59) & 0.96 (102/106) \\
	$21-30$ & 0.93 (471/504) & 0.96 (684/714) \\
	$31-40$ & 0.97 (373/383) & 0.92 (226/245) \\
	$41-50$ & 0.98 (171/174) & 0.94 (84/89) \\
	$51-60$ & 0.98 (236/240) & 0.90 (84/93) \\
	$61-70$ & 0.98 (122/125) & 0.91 (49/54) \\
	$>$ 70 & 0.95 (86/91) & 0.73 (61/84) \\
	\hline
	Overall accuracy & \textbf{0.93 (1723/1855)} & \textbf{0.91 (1544/1701)} \\ 
	\bottomrule
	\end{tabular}
\end{center}
\vspace{1mm}
{\small 

*: (M/N) denotes the correctly classified instances M over the total instances N for a particular gender category and age group.}
\end{table}

\subsubsection{FR (Face Recognition) Score} \label{sec:facenet}
 
We institute a second quantitative measure to evaluate the identity-preserving effect in the age-progressed photos. To this end, we utilize a pre-trained FaceNet classifier\cite{schroff2015facenet}, a state-of-the-art face recognition system developed by Google in 2015. The system achieved record-high accuracy on a range of face recognition benchmark datasets, including $99.63\%$ accuracy in the Wild (LFW) dataset\cite{huang2008labeled} for face verification. Conceptually, FaceNet learns a direct mapping from face images to a compact Euclidean space where distances correspond to a measure of face similarity.

 \begin{figure*}[!h]
\centering
\includegraphics[trim = 80 0  50 0, clip, scale=0.95]{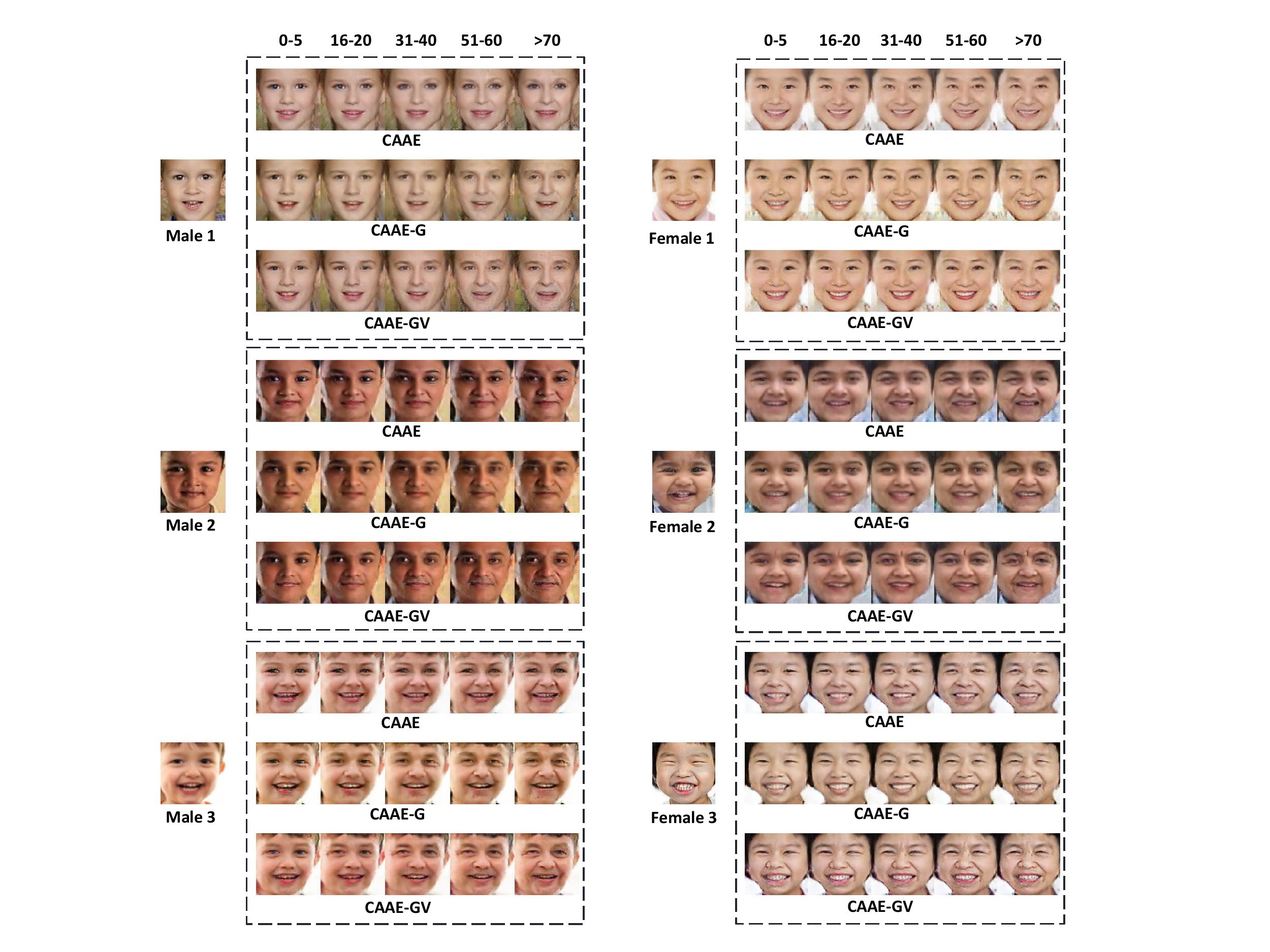}
\caption{Visual Assessment of Age-progressed Photos Across Three Models. CAAE-G denotes a modified CAAE model with additional gender information. CAAE-GV denotes our model, a modified CAAE model with both gender and VGG components.}
\label{fig:faces}
\end{figure*}

We define the FR score for a particular age group to be the accuracy of applying the FaceNet classifier to all generated and corresponding input image pairs within the group. A higher FR score indicates an overall higher likelihood of being the same person; thus, it can assess the identity-preserving effects of age-progressed photos. Clearly, the FaceNet model's performance is dependent on the distance threshold that the classifier uses to decide if two faces are considered from the same person.

\renewcommand{\arraystretch}{1.4}
\begin{table*}[!t] 
\caption{Gender Score Comparison for Four Models Across Ten Age Groups}
\label{tab:male}
\small
\begin{center}
		\begin{tabular}{c||c|l|l|l||c|l|l|l}
		
	\toprule
	\multirow{2}{*}{Age Group} &\multicolumn{4}{c||}{Male Samples} & \multicolumn{4}{|c}{Female Samples}\\\cline{2-9}
	&CAAE&CAAE-G &CAAE-V&CAAE-GV &CAAE&CAAE-G &CAAE-V&CAAE-GV\\
	\hline
	$0-5$ & $0.65$ & $0.77$ (19.5\%)* & $0.65$ (-0.1\%) & $0.68$ (5.2\%) & $0.61$ & $0.66$ (8.0\%) & $0.64$ (5.3\%) & $0.70$ (14.7\%)\\
	$6-10$ & $0.53$ & $0.74$ (40.3\%) & $0.54$ (3.5\%) & $0.66$ (25.6\%) & $0.68$ & $0.73$ (7.1\%) & $0.71$ (3.9\%) & $0.79$ (15.0\%) \\
	$11-15$ & $0.44$ & $0.71$ (61.2\%) & $0.43$ (-2.5\%) & $0.61$ (38.2\%) & $0.75$ & $0.79$ (5.6\%) & $0.77$ (3.6\%) & $0.83$ (11.0\%)\\
	$16-20$ & $0.34$ & $0.78$ (129.4\%) & $0.36$ (6.1\%) & $0.68$ (100.6\%) & $0.81$ & $0.81$ (0.1\%) & $0.81$ (-0.1\%) & $0.87$ (6.4\%)\\
	$21-30$ & $0.31$ & $0.78$ (152.2\%) & $0.28$ (-7.6\%) & $0.70$ (126.4\%) & $0.83$ & $0.87$ (6.0\%) & $0.84$ (2.0\%) & $0.91$ (10.2\%) \\
	$31-40$ & $0.36$ & $0.84$ (133.9\%) & $0.35$ (-2.3\%) & $0.76$ (113.1\%) & $0.80$ & $0.86$ (7.1\%) & $0.80$ (0.2\%) & $0.91$ (13.5\%)\\
	$41-50$ & $0.42$ & $0.89$ (110.2\%) & $0.40$ (-6.8\%) & $0.83$ (95.4\%) & $0.76$ & $0.83$ (9.5\%) & $0.77$ (1.5\%) & $0.88$ (16.0\%) \\
	$51-60$ & $0.44$ & $0.91$ (106.1\%) & $0.42$ (-5.6\%) & $0.86$ (93.2\%) & $0.73$ & $0.81$ (11.6\%) & $0.77$ (5.1\%) & $0.89$ (21.3\%) \\
	$61-70$ & $0.45$ & $0.90$ (101.2\%) & $0.43$ (-4.3\%) & $0.85$ (89.7\%) & $0.71$ & $0.80$ (11.8\%) & $0.76$ (6.0\%) & $0.85$ (18.8\%)\\
	$>$70 & $0.47$ & $0.92$ (94.0\%) & $0.45$ (-5.9\%) & $0.87$ (83.0\%) & $0.70$ & $0.68$ (-3.1\%) & $0.75$ (7.5\%) & $0.77$ (10.9\%)\\
	\hline
	Average & $0.44$ & \bf 0.82 (94.8\%) & $0.43$ (-2.5\%) & \bf 0.75 (77.0\%) & $0.74$ & $0.78$ (6.4\%) & $0.76$ (3.5\%) & \bf 0.84 (13.8\%)\\
	\bottomrule
	\end{tabular} 
\end{center}
\vspace{1mm}
	{\small
\hspace*{1cm} *: Numbers in parenthesis are the percentage gains over the CAAE model \\
\hspace*{1cm}  CAAE-G: CAAE model + gender component\\
\hspace*{1cm} CAAE-V: CAAE model + VGG component\\
\hspace*{1cm} CAAE-GV: CAAE model + both gender and VGG components

}
\end{table*}

\subsection{Experimental Results}

Our model consists of two extensions to the baseline CAAE model, each of which serves a different purpose. To examine the individual and combined effects of these two components, we present and analyze the results of four models, CAAE, CAAE-G, CAAE-V, and CAAE-GV, which denote the baseline, baseline+gender, baseline+VGG, and baseline+gender+VGG respectively. Each model is trained using the UTKFace dataset and applied to 1156 male and 1207 female infant and toddler photos to generate simulated faces in ten age groups. 

Fig. \ref{fig:faces} presents sample simulated faces across different models and the age spectrum. We selected three test samples from each of the male and female categories. For each test photo, we display a simulation block of three rows, each of which consists of predicted faces by the indicated model for the 0-5, 16-20, 31-40, 51-60, and $>$70 age buckets. Note that the CAAE-V model was excluded in Fig. \ref{fig:faces}  because we found that the effects of adding the VGG component alone are not visually pronounced. Nevertheless, we analyze its impact quantitatively using the gender and FR scores.

\subsubsection{Visual Evaluation of the Gender Component}

We examine the visual effects of the augmented gender indicator by comparing the faces in the top two rows in each simulation block in Fig. \ref{fig:faces}. We observe that, for male samples, there is a pronounced difference in masculinity between the two rows of simulated faces. Furthermore, the discrepancies are more salient as age progresses. The baseline CAAE model (i.e., top row) tends to project the faces toward the female direction, and the CAAE-G model (i.e., second row) corrects the mistake and generates the faces along the correct gender trajectory. 

For the female samples, the difference in gender characteristics between the CAAE and CAAE-G models is not as notable as that of the male ones. This is expected if the baseline model is biased toward females. Nevertheless, we do observe the differences and it is arguable that the second rows exhibit more femininity compared to the first rows. To establish a more rigorous evaluation, we employ the "gender score" (described in Section \ref{sec:gs}) to quantify the fidelity of the sex category in the simulated photos compared to the ground-truth. 

The third row in each of the simulation blocks presents the results of the CAAE-GV model. We observe that the gender advantage of the CAAE-G model is preserved and the faces are more realistic due to the additional VGG components. We quantitatively confirm this observation in Section \ref{sec:vggquant}.

\subsubsection{Quantitative Evaluation of the Gender Component}

Tables \ref{tab:male}  present the gender scores of different models over the age spectrum for both male and female categories. For the male samples, we observe that the CAAE-G model demonstrates an average gain of 94.8\% across all age groups. This confirms our visual observation discussed in the above section. We further observe that the improvements are consistent across all age groups and there is a monotonic upward trend until the 31-40 group. The trend then reverts itself afterwards to decrease monotonically. There is a sharp gain (i.e., from 61.2\% to 129.4\%) when entering the 16-20 group. This is consistent with physical development during adolescence when gender features become apparent in facial images.    

For the female category, we observe that the CAAE-G model demonstrates an overall 6.4\% gain for the female test samples over the baseline model. Compared to the male case, the gains are consistent but at a much smaller scale.  They are also more evenly distributed over the age spectrum. We conclude that the additional gender information also helps to produce more sex consistent photos in female samples in all age groups. Furthermore, compared to the male category, the considerably higher gender scores in the CAAE column for the female samples suggests that the baseline model has a bias towards the female gender in its simulations.   

The CAAE-V columns in Table \ref{tab:male} demonstrate that the VGG component does not have a high influence on the gender characteristics in the simulated faces. This is plausible because the purpose of the VGG discriminator is for preserving the facial features that are essential for identity verification. In our study, all input photos are from infants and toddlers whose gender features are not fully developed on their faces. Indeed, the CAAE-V model exhibits a marginal improvement (3.5\%) for the female class and a marginal negative impact (-2.5\%) for the male class over the baseline model. However, we present next that the VGG component leads to a significant improvement in preserving individual identities in the face aging simulations.

\renewcommand{\arraystretch}{1.4}
\begin{table}[!t] 
\caption{ Distance ($L_2$) Statistics Between Original and Simulated Faces}
 \label{tb:stats}
\begin{center}
	\begin{tabular}{r|c|l|l|l   }
	\toprule
	 &CAAE& CAAE-G & CAAE-V& CAAE-GV \\
	\hline
	min & $0.33$ & $0.29$ (10.8\%) & $0.23$ (31.6\%) & \bf {0.17 (49.8\%)} \\
	max & $5.39$ & $4.61$ (14.5\%) & $4.56$ (15.4\%) & \bf 4.81 (10.7\%)  \\ 
	mean & $1.88$ & $1.83$ (2.3\%) & $1.86$ (0.8\%) & \bf{1.77 (5.8\%)} \\ 
	SD & $0.70$ &$0.73$ (-3.1\%) & $0.73$ (-3.1\%) & $0.76$ (-8.2\%) \\
	10-PCTL & $1.02$ & $0.96$ (6.3\%) & 0.98 (4.0\%) & \bf {0.85 (16.8\%) }  \\
	20-PCTL & $1.24$ & $1.19$ (4.4\%) & $1.19$ (4.4\%) & \bf{1.07 (14.1\%)}  \\
	30-PCTL & $1.45$ & $1.37$ (5.3\%) & $1.39$ (3.9\%) & \bf{1.27 (12.3\%)} \\
	40-PCTL & $1.63$ & $1.55$ (4.8\%) & $1.60$ (1.9\%) & \bf{1.47 (10.1\%)} \\
	50-PCTL & $1.81$ & $1.76$ (2.9\%) & $1.79$ (1.2\%) & \bf{1.68 (7.2\%)}  \\
	60-PCTL & $2.01$ & $1.96$ (2.3\%) & $2.00$ (0.4\%) & \bf{1.91 (4.9\%)}  \\
	70-PCTL & $2.20$ & $2.19$ (0.8\%) & $2.22$ (-0.7\%) & \bf{2.14 (3.0\%)} \\
	80-PCTL & $2.46$ & $2.45$ (0.4\%) & $2.48$ (-0.7\%) & \bf{2.44 (0.7\%)}  \\
	90-PCTL & $2.83$ & $2.83$ (0.2\%) & $2.86$ (-0.9\%) & \bf{2.79 (1.5\%)} \\
		\bottomrule
	\end{tabular}
	\vspace*{1mm}
\end{center}

*: Numbers in parenthesis are the percentage gains over the CAAE model
\end{table}

\subsubsection{Visual Evaluation of the VGG Component}
The effect of the VGG component can be visualized by comparing the second and third rows in each of the simulation blocks in Fig. \ref{fig:faces}. While both models have integrated the gender indicator, the CAAE-GV model (i.e., third row) has the extra VGG loss (described in Section \ref{sec:identity}) incorporated into its model training process. 

We observe that the third row has more vivid facial expressions and better skin tones in general compared to the second row. For all examples, faces in the third row also develop more natural wrinkles and nasolabial folds in the older groups. Consequently, the CAAE-GV model produces more realistic simulations compared to the CAAE-G model. We confirm our observation more rigorously using the "FR score" in the next section.

\subsubsection{Quantitative Evaluation of the VGG Component}
\label{sec:vggquant}

As described in Section \ref{sec:facenet}, we institute an "FR (Face Recognition) score" to quantify the degree of identity preservation throughout the face aging process. Because the FR score relies on the distance measure in the embedded Euclidean space, we first examine the statistics of the $L_2$ distance between the original image and the age-progressed faces for different models.

Table~\ref{tb:stats} presents, for each model, the distribution of distances between the test samples and their corresponding generated images. The smaller the distance, the more likely the two faces belong to the same person. We observe that the CAAE-GV model produces the smallest mean (1.77) among the four distributions with a 5.8\% gain over the baseline model, while adding gender information or the VGG component alone results in a 2.3\% and 0.8\% gain respectively. Similarly, for the min/max statistics, the CAAE-GV model demonstrates 49.8\%/10.7\% improvements over the baseline model, while the CAAE-G and CAAE-V models exhibit 10.8\%/14.5\% and 31.6\%/15.4\% gains respectively.
Furthermore, the advantage of the CAAE-GV model over the other three models is consistent across all percentile groups.

\/*
\renewcommand{\arraystretch}{1.4}
\begin{table*}[!t] 
\caption{ Distance ($L_2$) Statistics Between Original and Simulated Faces}
 \label{tb:stats}
\begin{center}
	\begin{tabular}{r|c|l|l|l  ||c|l|l|l }
	\toprule
	&\multicolumn{4}{|c||} {$L_2$ Distance} & \multicolumn{4}{|c} {Cosine Similarity} \\
\hline
	 &CAAE& CAAE-G & CAAE-V& CAAE-GV & CAAE& CAAE-G & CAAE-V& CAAE-GV\\
	\hline
	min & $0.33$ & $0.29$ (10.8\%) & $0.23$ (31.6\%) & \bf {0.17 (49.8\%)} &0.127& 0.124 (2.7\%)& 0.102 (19.8\%)& 0.101 (20.5\%)\\
	max & $5.39$ & $4.61$ (14.5\%) & $4.56$ (15.4\%) & \bf 4.81 (10.7\%)  &0.616& 0.553 (10.3\%)& 0.552 (10.5\%)& 0.563 (8.68\%)\\
	mean & $1.88$ & $1.83$ (2.3\%) & $1.86$ (0.8\%) & \bf{1.77 (5.8\%)} &0.326 & 0.320 (1.8\%) & 0.324 (0.72\%)& 0.314 (3.75\%)\\
	SD & $0.70$ &{\color{red}$0.73$ (-3.1\%) }& {\color{red}$0.53$ (-3.1\%) }& $0.58$ (-8.2\%) &0.068 & 0.070 (-3.0\%)& 0.070 (-3.0\%)& 0.075 (-10.36\%)\\
	10-PCTL & $1.02$ & $0.96$ (6.3\%) & 0.98 (4.0\%) & \bf {0.85 (16.8\%) }  &0.241& 0.230 (4.6\%) & 0.234 (2.9\%)& 0.217 (10.0\%)\\
	20-PCTL & $1.24$ & $1.19$ (4.4\%) & $1.19$ (4.4\%) & \bf{1.07 (14.1\%)}  &0.266& 0.256 (3.8\%)& 0.261 (1.88\%)& 0.244 (8.27\%)\\
	30-PCTL & $1.45$ & $1.37$ (5.3\%) & $1.39$ (3.9\%) & \bf{1.27 (12.3\%)} &0.287&0.278 (3.1\%)& 0.283 (1.39\%)& 0.268 (6.62\%) \\
	40-PCTL & $1.63$ & $1.55$ (4.8\%) & $1.60$ (1.9\%) & \bf{1.47 (10.1\%)} &0.306& 0.299 (2.3\%)& 0.304 (0.65\%) & 0.290 (5.23\%)\\
	50-PCTL & $1.81$ & $1.76$ (2.9\%) & $1.79$ (1.2\%) & \bf{1.68 (7.2\%)}  &0.324& 0.318 (1.9\%)& 0.323 (0.3\%)& 0.311 (4.01\%)\\
	60-PCTL & $2.01$ & $1.96$ (2.3\%) & $2.00$ (0.4\%) & \bf{1.91 (4.9\%)}  &0.344& 0.337 (2.0\%)& 0.341 (0.9\%)& 0.334 (2.91\%)\\
	70-PCTL & $2.20$ & $2.19$ (0.8\%) & $2.22$ (-0.7\%) & \bf{2.14 (3.0\%)} &0.363& 0.359 (1.1\%)& 0.361 (0.6\%)& 0.355 (2.20\%)\\
	80-PCTL & $2.46$ & $2.45$ (0.4\%) & $2.48$ (-0.7\%) & \bf{2.44 (0.7\%)}  &0.385& 0.383 (0.5\%)& 0.385 (0.00\%)& 0.383 (0.52\%)\\
	90-PCTL & $2.83$ & $2.83$ (0.2\%) & $2.86$ (-0.9\%) & \bf{2.79 (1.5\%)} &0.413& 0.413 (0.00\%)& 0.419 (-1.45\%)& 0.411 (0.48\%)\\
	\bottomrule
	\end{tabular}
	\vspace*{1mm}
\end{center}

*: Numbers in parenthesis are the percentage gains over the CAAE model
\end{table*}

*/


Next, we compare the FR score of the four models. Recall that an FR score represents the FaceNet model's classification accuracy in determining if the simulated face is the same as the original person's. This accuracy depends on the distance threshold we use to make the verification decision. A lower threshold (i.e., closer distance) requires a higher similarity between the paired images to be identified as the same person. On the other hand, a higher threshold relaxes the verification criteria. In the extreme case, when the threshold is set above the max distance of the distributions in Table \ref{tb:stats}, all FR scores will degenerate to 1. Thus, the FR score improvement with a lower threshold is more substantial than that with a higher threshold, even though lower thresholds are associated with lower FR scores.

Table~\ref{tb:fr_score} presents the FR score for different models using three distance thresholds. We observe that in each model's corresponding column, the FR score increases as expected when we relax the identity verification criteria. When we set a high verification standard (threshold=1.6), the CAAE-GV model exhibits a 22.4\% gain over the baseline CAAE model. The CAAE-G and CAAE-V models offer a 12.9\% and 5.3\% gain, respectively. As we lower the criteria (i.e., increase the threshold), the advantage of our model reduces due to the expected diminishing FR score differences among all four models.

\renewcommand{\arraystretch}{1.4}
\begin{table}[!t] 
\caption{Face Recognition (FR) Score Comparison of Four Models}
\label{tb:fr_score}
	\begin{tabular}{c|c|c|c|c}
	\hline
	Threshold & CAAE & CAAE-G & CAAE-V & CAAE-GV \\
	\hline
	\bf 1.6 & \bf 0.38 & {\bf 0.43 (12.9\%)*} & \bf{0.40 (5.3\%)} & \bf{0.47 (22.4\%)} \\
	$2.0$ & $0.60$ & $0.62$ (3.9\%) & $0.60$ (0.8\%) & $0.65$ (8.2\%) \\
	$2.5$ & $0.82$ & $0.82$ (-0.2\%) & $0.81$ (-1.0\%) & $0.82$ (0.3\%) \\
	\hline
	\end{tabular} 
	\vspace*{1mm}

*: Numbers in parenthesis are the percentage gains over the CAAE model
\end{table}

\section{Conclusion}
\label{sec:conclusion}
In this paper, we focused on the challenging task of age progression/regression for photos of infants and toddlers. 
We proposed two enhancements to the existing CAAE architecture to help ensure gender and identity consistencies in the face aging process. In our approach, we augmented the input vector of the generator with gender information since young faces lack salient gender characteristics. We further strengthened the model with a new identity preservation component based on facial features extracted by the VGG19 convolutional neural network. Our experimental results demonstrate significant visual and quantitative improvements over the CAAE model for our particular task. Our methods and findings can be adopted by other deep learning approaches in the face progression and regression studies.

\bibliographystyle{IEEEtran}
\bibliography{ref}

\end{document}